\documentclass[conference,10pt]{IEEEtran}
\IEEEoverridecommandlockouts
\usepackage{cite}
\usepackage[colorlinks, linkcolor=magenta, anchorcolor=green, citecolor=blue]{hyperref}
\usepackage{amsmath,amssymb,amsfonts}
\usepackage[numbers,sort&compress]{natbib}
\usepackage{algorithmic,algorithm}
\usepackage{fancyhdr}
\usepackage{graphicx}
\usepackage{textcomp}
\usepackage{xcolor}
\def\BibTeX{{\rm B\kern-.05em{\sc i\kern-.025em b}\kern-.08em
    T\kern-.1667em\lower.7ex\hbox{E}\kern-.125emX}}
\usepackage{multirow}
\usepackage{makecell}
\usepackage{array}

\usepackage{pifont}
\iftrue
\fi

\usepackage{etoolbox}
\apptocmd{\thebibliography}{\scriptsize}{}{}

\begin{document}

\title{\huge MegaGraph: Towards Efficient Training of Large-Scale Graph Transformers with Automated Hybrid Parallelism}

\author{
    \IEEEauthorblockN{
        Tong Qiao\IEEEauthorrefmark{1},
        Ao Zhou\IEEEauthorrefmark{2},
        Yingjie Qi\IEEEauthorrefmark{1},
        Chunming Hu\IEEEauthorrefmark{2}\IEEEauthorrefmark{3} and
        Jianlei Yang\IEEEauthorrefmark{1}\IEEEauthorrefmark{4}\IEEEauthorrefmark{3}
    }
    \IEEEauthorblockA{
        \IEEEauthorrefmark{1}School of Computer Science and Engineering, Beihang University, China \\
        \IEEEauthorrefmark{4}Qingdao Research Institute, Beihang University, Qingdao, China \\
        \IEEEauthorrefmark{2}School of Software, Beihang University, China\\
        \IEEEauthorrefmark{3}State Key Laboratory of Complex and Critical Software Environment, Beihang University, China
    }
    \thanks{This work was supported in part by the National Key R\&D Program of China (Grant No. 2024YFB4505601), the Shandong Provincial Natural Science Foundation (Grant No. ZR2026LJS009) and the Beijing Natural Science Foundation (Grant No. L243031). The corresponding authors are \textit{Jianlei Yang} and \textit{Chunming Hu}. Email: \url{jianlei@buaa.edu.cn}, \url{hucm@buaa.edu.cn}.}
}

\maketitle
\thispagestyle{plain}
\pagestyle{plain}

\bstctlcite{IEEEexample:BSTcontrol}

\begin{abstract}
Graph Transformers (GTs) offer superior representation capabilities
by overcoming the depth limitations and over-smoothing issues
of traditional Graph Neural Networks (GNNs).
However, scaling GTs to large graphs poses critical bottlenecks.
Specifically, the attention score matrix and its associated topology-aware bias matrix jointly incur significant per-layer memory overhead,
and heavy graph embedding layers result in severe workload imbalances.
These characteristics are unique to GT training and are not addressed
by parallelism techniques designed for either conventional GNNs or Transformers,
making a dedicated solution necessary.
This paper introduces MegaGraph,
the first automated hybrid parallelism framework designed for efficient GT training.
MegaGraph designs three specialized strategies,
namely graph-aware context parallelism,
heterogeneous pipeline parallelism,
and hybrid data parallelism,
to support efficient training on large-scale graphs.
However, coordinating these three parallelism strategies
yields an exponentially large configuration space.
To address this complexity,
an automatic search engine leverages precise cost models
via a \textit{Profile $\to$ Model $\to$ Search} workflow
to identify the optimal parallelism configuration.
Evaluations demonstrate that MegaGraph enables training on large-scale graphs
where state-of-the-art baselines fail due to out-of-memory (OOM) errors.
The framework reduces per-device peak memory by up to 77.8\%
and achieves up to 4.51$\times$ training speedup
while maintaining model accuracy.
\end{abstract}


\section{Introduction}
\label{sec:intro}

Graph Neural Networks (GNNs) have demonstrated significant success
across diverse domains,
such as recommendation systems~\cite{yang2023dgrec},
link prediction~\cite{wang2024efficient},
and knowledge graph analysis~\cite{bi2023bridged}.
Traditional GNNs rely on iterative local aggregation mechanisms
to handle the non-Euclidean nature of graph data~\cite{kipf2017semi, gilmer2017neural}.
However, this mechanism faces a fundamental dilemma.
Stacking layers to expand the receptive field
inevitably triggers over-smoothing,
which severely degrades the representation capability
of deep architectures~\cite{li2018deeper, chen2020measuring}.
Recently, the Transformer architecture~\cite{vaswani2017attention}
has driven significant advances in Large Language Models (LLMs),
motivating researchers to extend it to graph-structured data.
By incorporating structural encoding into the attention mechanism,
Graph Transformers (GTs) leverage global attention to connect all nodes,
breaking the depth limitations of traditional GNNs~\cite{ying2021transformers, rampasek2022graphgps, wu2022nodeformer, zhang2022hierarchical}.

\begin{figure}[t]
    \centering
    \includegraphics[width=1\columnwidth]{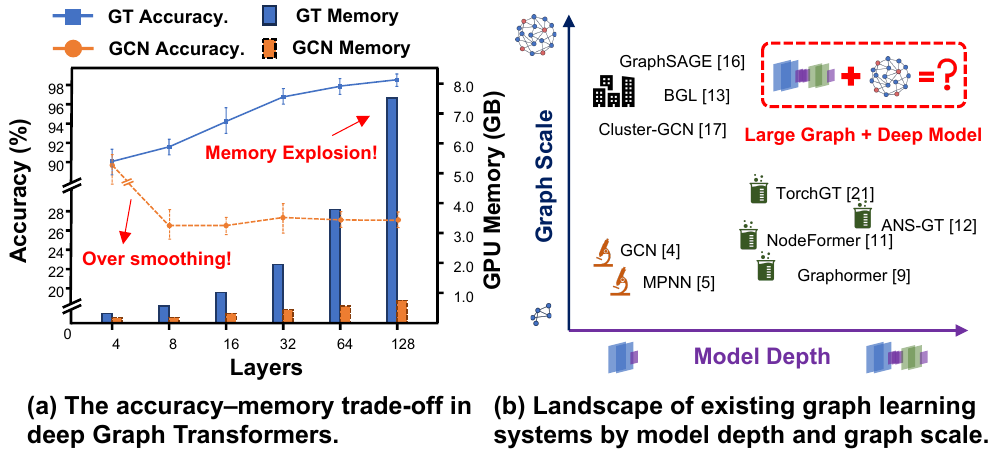}
    \vspace{-16pt}
    \caption{Scalability of graph representation learning.}
    \label{fig:motivation}
\end{figure}

However, the global attention mechanism incurs an O($N^2$) per-layer memory footprint,
and the total memory further accumulates as the number of layers increases.
As shown in Fig.~\ref{fig:motivation}(a),
GT accuracy improves steadily with increasing depth,
overcoming the over-smoothing issue of GNNs,
but GPU memory grows rapidly as well.
The conflict between the expressive power of deep GTs
and the memory ceiling of a single GPU
constitutes the core bottleneck,
making distributed training essential.

Furthermore, existing parallelism techniques cannot be directly applied to GTs.
LLM parallelism strategies assume dense and regular computational workloads,
but graph structure violates these assumptions.
The attention bias matrix loses row-column alignment
during ring attention rotation in standard context parallelism (CP).
In addition, the node embedding lookup table scales linearly with the total node count,
breaking the uniform partitioning assumption of standard pipeline parallelism (PP).
GNN distributed training systems~\cite{liu2023bgl} are designed for shallow message-passing architectures,
relying on neighbor sampling and graph partitioning for data parallelism.
These systems lack support for the attention partitioning
and pipeline scheduling required by deep GTs.
Moreover, the configuration space for hybrid parallelism
grows combinatorially, making manual tuning intractable.
Although adaptive and automatic GNN training on CPU-GPU
heterogeneous platforms has been explored~\cite{qiao2024gnnavigator, qiao2025towards},
these approaches target conventional message-passing GNNs
rather than the deep GTs we study.

As shown in Fig.~\ref{fig:motivation}(b),
existing graph learning systems fall into three categories.
Classical full-graph aggregation GNNs with shallow architectures
are primarily designed for small and medium-scale graphs.
Sampling-based mini-batch GNNs~\cite{hamilton2017inductive, chiang2019cluster} can scale to large graphs,
but remain constrained by the over-smoothing problem,
preventing the design of deeper network architectures.
Current GT models leverage global attention to support deeper architectures,
but can only handle small and medium-scale graphs
and struggle to scale to large-scale graphs.
A clear gap remains at the intersection of large-scale graphs and deep models.

We propose MegaGraph, the first automated hybrid parallelism
framework designed for GT training, which supports training on large-scale graphs with deep network architectures.
MegaGraph comprises two core modules:
a hybrid parallelism backend tailored to Graph Transformer workloads,
and an automatic search engine that identifies the optimal strategy
from the combinatorially growing configuration space.
Our main contributions are summarized as follows.
\begin{itemize}
    \item We propose the MegaGraph parallelism backend
    with three specialized strategies.
    Graph-aware context parallelism distributes both the attention computation
    and the associated bias matrix across multiple GPUs.
    Heterogeneous pipeline parallelism balances workloads across stages
    through non-uniform layer partitioning.
    Hybrid data parallelism further scales training throughput
    along the graph data dimension.
    \item We design an automatic search engine
    following a \textit{Profile $\to$ Model $\to$ Search} workflow,
    which leverages precise cost models and memory-aware pruning
    to identify the optimal parallelism configuration
    within seconds from an exponentially large search space.
    \item Evaluations demonstrate that MegaGraph enables training
    on large-scale graphs where existing baselines fail with OOM,
    reducing peak memory by 77.8\% and achieving 4.51$\times$ training speedup
    compared with the best-performing baseline, while maintaining model accuracy.
\end{itemize}

\section{Preliminaries \& Motivations}

\subsection{Preliminaries: Graph Transformer}

\begin{figure}[t]
    \centering
    \includegraphics[width=1\columnwidth]{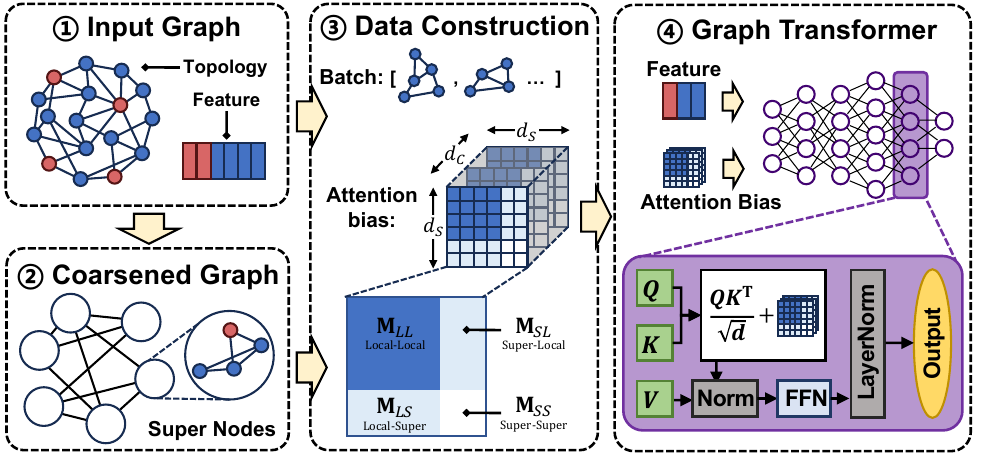}
    \vspace{-15pt}
    \caption{The hierarchical Graph Transformer training workflow
    comprising graph coarsening, data construction, and Transformer layer computation.}
    \label{fig:gt}
\end{figure}

Fig.~\ref{fig:gt} illustrates the training workflow of hierarchical GTs~\cite{zhang2022hierarchical}.
Given an input graph with topology and node features,
graph coarsening first compresses it into a set of super nodes,
each containing an internal subgraph.
During training, nodes are sampled to form mini-batches.
The attention bias matrix $\mathbf{b}_{attn}$ is then constructed
based on the topological relationships
among local nodes within the batch and between local nodes and super nodes.
This mechanism is known as structural encoding,
which encodes topological relationships as additive bias terms
injected into the multi-head attention computation.
In each Transformer layer,
node features are linearly projected to produce
the query matrix $\mathbf{Q}$, key matrix $\mathbf{K}$, and value matrix $\mathbf{V}$.
The attention is computed as:
\begin{equation}
    \text{Attention}(\mathbf{Q}, \mathbf{K}, \mathbf{V}) =
    \text{Softmax}\left(\frac{\mathbf{Q}\mathbf{K}^\top}{\sqrt{d}}
    + \mathbf{b}_{attn}\right)\mathbf{V}.
\end{equation}
Through graph coarsening, the global attention complexity is reduced
from O($|\mathcal{V}|^2$) to O($N^2$),
where $N$ is the total number of nodes in a single batch,
including both local nodes and super nodes ($N \ll |\mathcal{V}|$).

\subsection{Motivations}

\textbf{Observation 1:}
Per-layer attention and bias matrices dominate GPU memory.
Even after coarsening,
each layer of a hierarchical GT must store
an $N \times N$ attention score matrix
and an equally sized $\mathbf{b}_{attn}$ bias matrix.
In standard context parallelism,
the sequence is evenly partitioned across GPUs,
where each GPU holds a portion of Q, K, and V,
and KV blocks are rotated via ring attention~\cite{liu2023ringattention} to complete the full attention computation.
However, this mechanism cannot be directly applied to GTs.
This mechanism assumes that attention scores are solely determined by $\mathbf{Q}\mathbf{K}^\top$,
so KV blocks can be correctly computed on any GPU after rotation.
In GTs, however, the attention computation additionally requires
the precomputed $\mathbf{b}_{attn}$,
which encodes topological relationships between nodes
and cannot be derived online from $\mathbf{Q}\mathbf{K}^\top$.
When KV blocks are rotated to another GPU,
the locally stored column partition of $\mathbf{b}_{attn}$
no longer corresponds to the rotated KV block,
leading to incorrect attention score computation.

The graph partitioning and neighbor sampling methods
commonly used in distributed GNN training are also inapplicable.
GNNs rely on neighbor aggregation~\cite{hamilton2017inductive},
where each node only accesses its local neighborhood,
allowing the graph to be partitioned into independent subgraphs for separate processing.
GT attention, in contrast, is all-to-all,
requiring each node to compute attention scores with every other node in the batch.
Partitioning into subgraphs would break the integrity of global attention.
Therefore, neither the context parallelism from LLM training
nor the graph partitioning and sampling from GNN training
can be directly applied to distribute the attention computation of GTs.

\textbf{Motivation 1:}
A new context parallelism strategy is needed
that jointly partitions the attention computation and the bias matrix across GPUs
while preserving their consistency at every computation step.

\begin{figure*}[t]
    \centering
    \includegraphics[width=\textwidth]{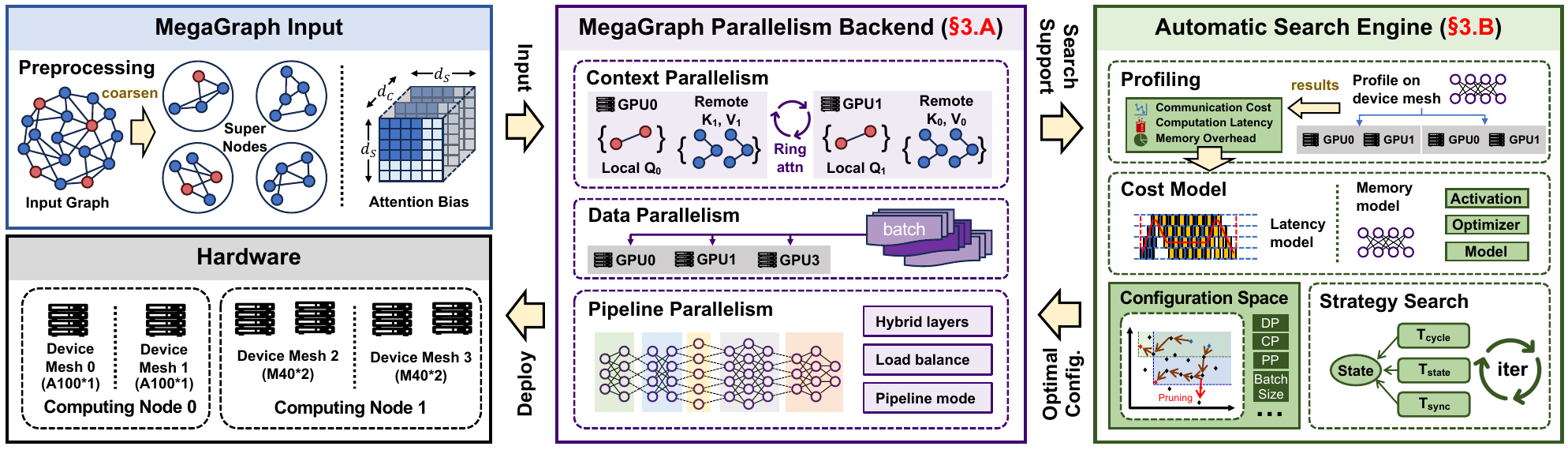}
    \vspace{-15pt}
    \caption{The overview of MegaGraph framework.}
    \label{fig:overview}
    \vspace{-15pt}
\end{figure*}

\textbf{Observation 2:}
Graph embedding and data construction introduce significant additional memory overhead.
Beyond the per-layer attention overhead discussed above,
the overall memory pressure also grows with model depth
as parameters, activations, and gradients accumulate across layers.
In LLM training, pipeline parallelism was proposed to address this problem
by distributing model layers across multiple GPUs~\cite{huang2019gpipe}.
Standard pipeline parallelism adopts uniform layer partitioning,
as vocabulary embedding tables and data preprocessing have small memory footprints,
resulting in approximately balanced workloads across stages.
In GT training, however,
the node embedding lookup table scales linearly with the total graph node count
as $|\mathcal{V}| \times d$,
consuming significantly more memory than a single Transformer layer's weights on large graphs.
In addition, the data construction module generates
attention bias tensors and feature buffers for each batch during training,
further increasing memory consumption.
These overheads are concentrated in the first stage of the pipeline,
causing uniform layer partitioning to result in
the first stage consuming far more memory than the remaining stages,
leading to severe workload imbalance and pipeline bubbles.

\textbf{Motivation 2:}
GT training demands a non-uniform layer partitioning strategy
that accounts for the varying memory overhead across pipeline stages.

\textbf{Observation 3:}
The hybrid parallelism configuration space grows combinatorially.
As discussed above, GT training requires both context parallelism and non-uniform pipeline parallelism. 
Context parallelism introduces a search dimension $P_{cp}$, while non-uniform pipeline parallelism must choose one of $\binom{L-1}{P_{pp}-1}$ layer partitions across $P_{pp}$ stages.
For example, $\binom{23}{3}=1771$ schemes for $L=24$, $P_{pp}=4$. Multiplying in the data-parallelism degree $P_{dp}$, the full configuration space grows combinatorially, making exhaustive search infeasible.

\textbf{Motivation 3:}
The optimal training configuration must be efficiently identified
from the combinatorially growing configuration space.

\vspace{-5pt}
\section{Methodology}

Fig.~\ref{fig:overview} illustrates the overall architecture of the MegaGraph framework.
This framework is logically decoupled into two core modules:
\begin{enumerate}
    \item \textbf{MegaGraph Parallelism Backend}:
    This module executes distributed training
    with three specialized parallelism strategies,
    each targeting a different granularity of the training workload.

    \item \textbf{Automatic Search Engine}:
    This module serves as the decision center of the framework.
    It predicts execution costs based on precise cost models.
    Following a \textit{Profile $\to$ Model $\to$ Search} workflow,
    this engine automatically identifies the optimal parallelism configuration
    for a given graph and hardware setup.
\end{enumerate}

These two modules interact closely and depend on each other.
First, the automatic search engine constructs precise mathematical cost models
using the profiling data provided by the parallelism backend.
Next, the automatic search engine determines the optimal parallelism strategy
within the vast configuration space based on these cost models.
Finally, this optimal strategy is deployed back to the parallelism backend
to guide distributed training.

\vspace{-3pt}
\subsection{MegaGraph Parallelism Backend}

We design three specialized parallelism strategies
and integrate them into the parallelism backend:
graph-aware context parallelism partitions memory along the attention dimension,
heterogeneous pipeline parallelism balances workloads along the layer dimension,
and hybrid data parallelism scales throughput along the graph data dimension.

\subsubsection{Graph-Aware Context Parallelism}

\begin{figure}[t]
    \centering
    \includegraphics[width=1\columnwidth]{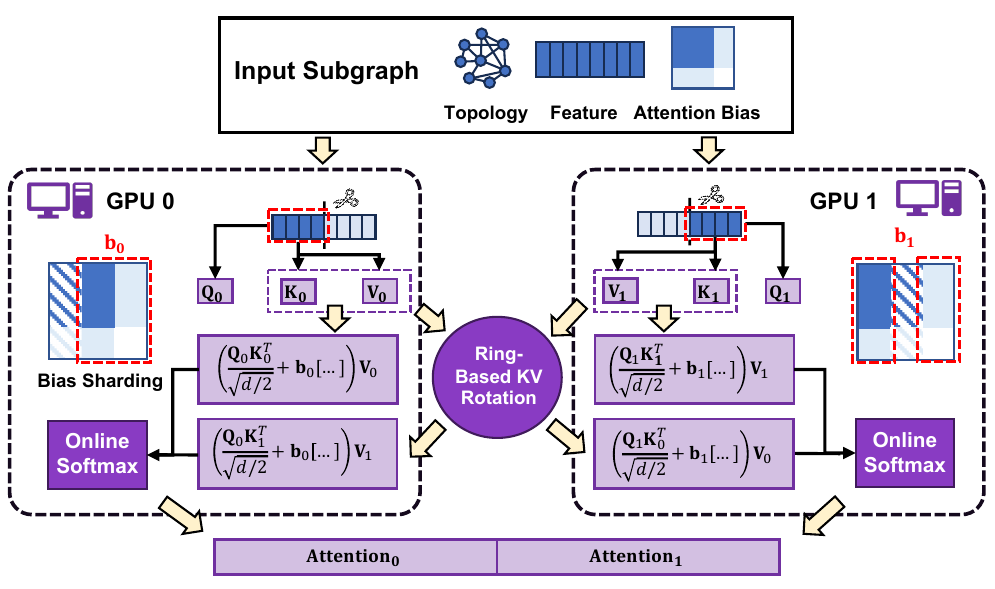}
    \vspace{-15pt}
    \caption{Graph-aware context parallelism: KV blocks rotate through the ring while bias columns are dynamically sliced to match each incoming KV block.}
    \label{fig:cp}
\end{figure}

We propose graph-aware context parallelism (GA-CP),
which distributes the attention computation and the bias matrix $\mathbf{b}_{attn}$
across multiple GPUs while preserving computational correctness.
GA-CP simultaneously partitions
both the attention computation and the bias matrix along the node dimension,
ensuring each rank retains the complete structural information
of its local nodes relative to all nodes in the batch.
During ring attention rotation of KV blocks,
the corresponding columns of $\mathbf{b}_{attn}$ are dynamically sliced
to match the current KV block, ensuring computational correctness.
The implementation comprises three core mechanisms,
as shown in Fig.~\ref{fig:cp} and detailed below.

\raisebox{-0.05ex}{\scalebox{1.15}{\ding{172}}}~\textit{Ring-Based KV Block Rotation.}
Each input batch contains $N'$ local graph nodes
and $N_s$ global super nodes, with total node count $N = N' + N_s$.
To avoid constructing the full $N \times N$ attention bias matrix
on a single device, we decompose the computation into $P_{cp}$ iterative steps
along a communication ring, where $P_{cp}$ denotes the context parallelism degree.
During computation, the regular nodes are partitioned while the global super nodes are replicated.
Each rank $i$ statically retains a local query shard $\mathbf{Q}_i$
of shape $[\frac{N'}{P_{cp}} + N_s, d]$,
covering its assigned $\frac{N'}{P_{cp}}$ regular nodes and all $N_s$ super nodes.
Similarly, the key-value pairs $(\mathbf{K}, \mathbf{V})$
are partitioned into identically sized blocks and circulate across the devices.
At the end of every iteration, rank $i$ sends its current KV blocks
to the downstream rank
and simultaneously receives the next KV blocks from the upstream rank.
After $P_{cp}$ iterations, the local $\mathbf{Q}_i$
traverses all $\mathbf{K}$ and $\mathbf{V}$ blocks in the batch,
completing the global information aggregation.

\raisebox{-0.05ex}{\scalebox{1.15}{\ding{173}}}~\textit{Query-Dimension Bias Sharding.}
Graph Transformers rely on dense structural bias matrices,
such as shortest path distances, to encode topological information.
Since each rank must correctly index into $\mathbf{b}_{attn}$
regardless of which KV block arrives after rotation,
simply partitioning the bias matrix by columns, as in standard CP, is insufficient.
We therefore introduce query-dimension bias sharding during the data loading phase,
which both resolves the alignment mismatch and eliminates the memory redundancy
of replicating the full $N \times N$ bias matrix.
Each rank $i$ instantiates only a partial bias shard
of shape $[\frac{N'}{P_{cp}} + N_s, N' + N_s]$,
which contains the structural information of its local nodes
relative to all nodes in the batch.
During the ring attention computation,
whenever rank $i$ receives a key block $\mathbf{K}_j$
containing both local graph nodes and global super nodes,
the system executes a two-part dynamic slicing operation.
Specifically, it extracts the $\frac{N'}{P_{cp}}$ columns corresponding to rank $j$'s regular nodes,
and the $N_s$ columns corresponding to the global super nodes.
These two sliced segments are then concatenated into a sub-block
that matches the shape of $\mathbf{K}_j$ for the attention computation.
The computation is formulated as follows:
\begin{equation}
    {Score}_{local} = \mathbf{Q}_i \cdot \mathbf{K}_{j}^T
    + \mathbf{b}_{i}[:, {Indices}(\mathbf{K}_j)].
\end{equation}
This strategy approximately reduces the memory footprint of the bias matrix
by a factor of $P_{cp}$ while fully preserving the topological information encoded in $\mathbf{b}_{attn}$.

\raisebox{-0.05ex}{\scalebox{1.15}{\ding{174}}}~\textit{Online Softmax \& Communication.}
Unlike standard ring attention where scores depend solely on the current KV block,
GT attention scores additionally include a step-varying bias term
$\mathbf{b}_i[:, {Indices}(\mathbf{K}_j)]$ that changes as different KV blocks rotate in.
Directly applying standard online softmax~\cite{dao2022flashattention} would update
\texttt{running\_max} and \texttt{running\_sum} based on incomplete scores,
leading to incorrect results.
We therefore incorporate the dynamically extracted bias slice into the score
before each online softmax update,
ensuring that normalization statistics always reflect the full bias-augmented score.
This formulation guarantees numerical stability
and allows attention computation to be pipelined with KV block communication,
with $P_{cp} - 1$ P2P exchanges each transmitting $2S \times d$ elements
(one $\mathbf{K}$ block and one $\mathbf{V}$ block),
where $S = N'/P_{cp} + N_s$ denotes the per-rank node count.

\begin{figure}[t]
    \centering
    \includegraphics[width=1\columnwidth]{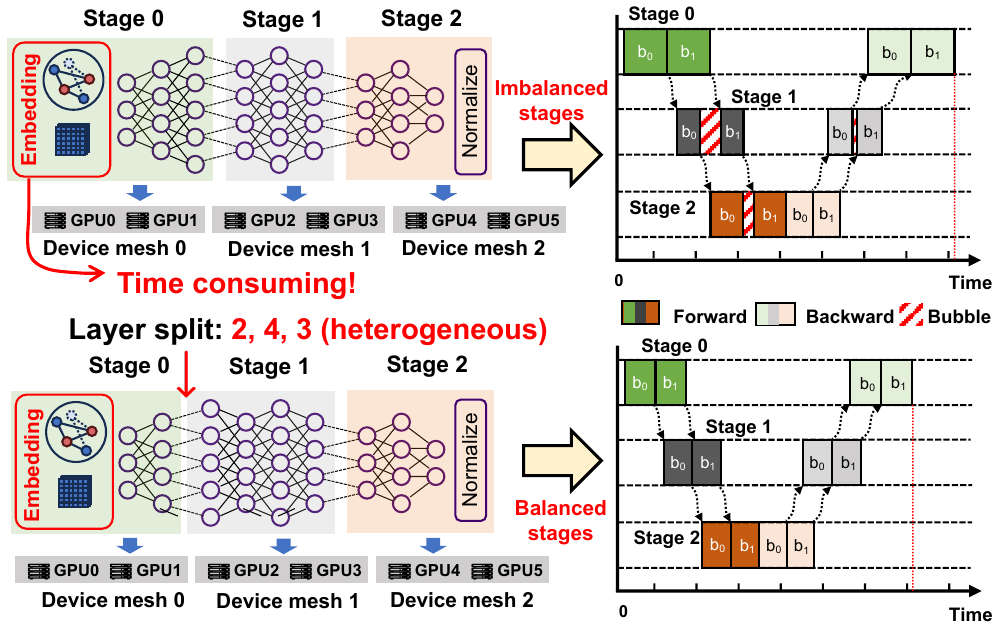}
    \vspace{-15pt}
    \caption{Comparison of pipeline parallelism strategies for Graph Transformers.
    Heterogeneous pipeline parallelism achieves better workload balance
    through non-uniform layer allocation.}
    \vspace{-8pt}
    \label{fig:pp}
\end{figure}

\subsubsection{Heterogeneous Pipeline Parallelism}

In a GT training pipeline,
the first stage hosts not only Transformer layers
but also the embedding lookup table and data construction module,
whose memory load far exceeds that of subsequent stages.
As shown in Fig.~\ref{fig:pp} (upper),
this imbalance under standard uniform layer partitioning
produces substantial pipeline bubbles at runtime,
wasting both memory and training time.

\begin{figure}[t]
    \centering
    \includegraphics[width=0.9\columnwidth]{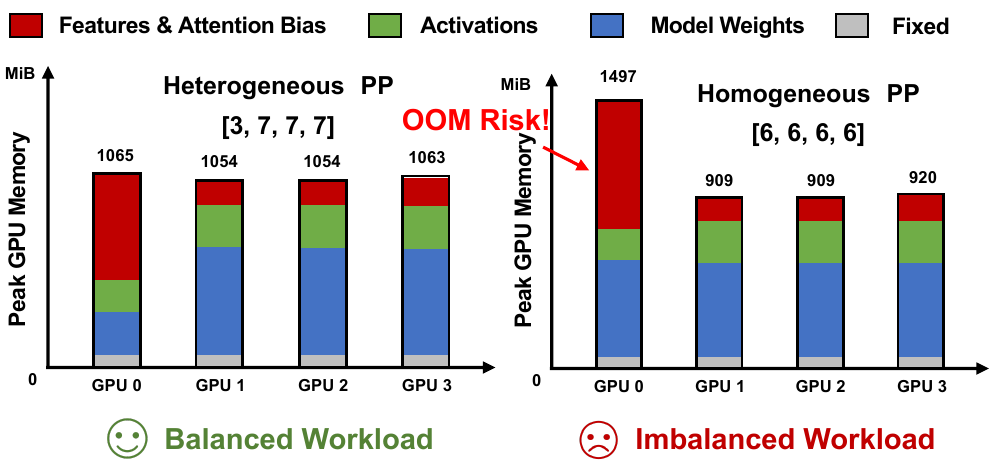}
    \vspace{-8pt}
    \caption{Peak memory footprint breakdown under different pipeline layer partition strategies.}
    \label{fig:pp-eval}
\end{figure}

MegaGraph introduces an embedding-aware non-uniform layer allocation strategy,
as illustrated in Fig.~\ref{fig:pp} (lower).
Rather than partitioning layers uniformly, MegaGraph explicitly accounts for
the disproportionate overhead of the embedding lookup table and data construction module
concentrated in the first stage,
assigning it fewer Transformer layers to compensate
and distributing the remaining layers more heavily across subsequent stages.
This brings per-stage execution times into balance,
directly reducing pipeline bubbles and improving throughput.
Execution follows the GPipe~\cite{huang2019gpipe} micro-batch schedule,
where each training batch is divided into $K$ micro-batches
that flow through the pipeline stages in sequence.
Fig.~\ref{fig:pp-eval} quantifies this effect.
When training a 24-layer GT across 4 GPUs with uniform partitioning,
the first stage consumes approximately 65\% more peak memory than subsequent stages.
Adjusting the allocation to $[3, 7, 7, 7]$ brings all stages to a comparable level.

\subsubsection{Hybrid Data Parallelism}

Beyond context and pipeline parallelism, data parallelism further scales training throughput along the graph data dimension.
The data construction module partitions mini-batches
across $P_{dp}$ device meshes,
where each mesh holds a complete model replica
and independently processes its data partition.
Gradients are synchronized across meshes via All-Reduce after each training step.

\subsection{Automatic Configuration Search}

Because our parallelism backend offers high flexibility,
finding the optimal configuration goes beyond simply setting the degrees of parallelism.
It requires fine-grained tuning,
such as specifying the exact number of layers assigned to each pipeline stage.
Consequently, traditional manual tuning methods based on empirical heuristics
are inadequate.
To identify the optimal parallelism configuration in large-scale graph training,
we design an automatic configuration search module
following a \textit{Profile $\to$ Model $\to$ Search} workflow,
as detailed in the following subsections.

\subsubsection{Profiling}

To control the runtime overhead of the automatic configuration search module,
we adopt a lightweight profiling method on actual hardware devices.
In the MegaGraph parallel architecture,
a device mesh consists of $P_{pp} \times P_{cp}$ GPUs
that host one complete model replica,
and the system comprises $P_{dp}$ such device meshes.
The effective computational capacity of each device mesh
is determined by the available computational power
and network communication bandwidth of its constituent GPUs,
which may vary due to existing workloads on shared environments.
We profile computation and communication in two independent phases.

During the computation profiling phase,
the system uses the hook functions provided by PyTorch.
We register hook functions on \texttt{torch.nn.Module},
which automatically intercept and record runtime timestamps
whenever these model components are invoked during the forward and backward passes.
For each candidate context parallelism degree $P_{cp}$,
the system runs a lightweight profiling pass and records
the forward propagation time $t_{fwd}$ and backward propagation time $t_{bwd}$
of a single Transformer layer,
along with the computation time of the node embedding layer $t_{emb}$
and the output projection layer $t_{proj}$,
since all these components are affected by the local node count $S$.

During the communication profiling phase,
the system benchmarks the network infrastructure of the GPU system.
We measure the actual network transmission time
for a given message size $\mathbf{C}$.
Specifically, the system records the point-to-point transfer time $t_{p2p}(\mathbf{C})$
used for tensor passing between pipeline stages.
Furthermore, the system evaluates the collective communication time $t_{allreduce}(\mathbf{C})$
used for global gradient synchronization within data parallel groups.

To ensure the profiling data reliably reflects actual performance,
the system first executes a few warm-up iterations to eliminate cold-start effects.
Subsequently, it collects data over 3 actual profiling runs
and calculates the average to produce reliable estimates.
This step filters out system jitter
and provides the expected runtime overhead that closely matches real-world scenarios.
These measurements are then used to parameterize the latency and memory cost models.

\subsubsection{Cost Model Construction}
\label{sec:costmodel-construction}

We develop memory and latency cost models
to evaluate the training overhead under different parallelism configurations.
Let $N = N' + N_s$ denote the total number of nodes in a batch.

\textbf{Memory cost model.}
The memory footprint of each device mesh $M_{mesh}$
consists of four components:
model weights $M_{weight}$ covering parameters, gradients, and optimizer states;
activations $M_{act}$ for intermediate tensors required by backward propagation;
features and bias $M_{feat}$ for attention bias tensors;
and fixed overhead $M_{fixed}$ for framework context, communication buffers,
and pipeline metadata.
The total memory is expressed as:
\begin{equation}
    M_{mesh} = M_{weight} + M_{act} + M_{feat} + M_{fixed}.
\end{equation}
The individual components are computed as follows.
For pipeline stage $i$ assigned $l$ Transformer layers,
the model weight memory is:
\begin{equation}
    M_{weight} = c_{fp} \times l \times (4d^2 + 2d \times d_{ff}) + M_{extra},
\end{equation}
where $d$ is the hidden dimension, $d_{ff}$ is the FFN dimension,
$c_{fp}$ is a constant that converts parameter count to memory bytes,
accounting for the storage of parameters, gradients, and optimizer states,
and $M_{extra}$ captures the additional parameters
of the embedding layer with input dimension $d_{in}$
and the projection layer with output dimension $d_{out}$:
\begin{equation}
    M_{extra} =
    \begin{cases}
    c_{fp} \times d_{in} \times d & \text{if } i = 0 \\
    c_{fp} \times d \times d_{out} & \text{if } i = P_{pp}-1 \\
    0 & \text{otherwise}
    \end{cases}.
\end{equation}
The activation memory depends on batch size $b$ and attention head count $h$,
while the feature-bias memory additionally accounts for the input feature buffer
at the first stage:
\begin{equation}
    M_{act} = l \cdot b \cdot \left(\frac{N'}{P_{cp}} + N_s\right)
    \cdot \left(c_{mlp} \cdot d + c_{attn} \cdot h \cdot (N' + N_s)\right),
\end{equation}
\begin{equation}
\begin{aligned}
M_{feat}
= b \cdot \left(\frac{N'}{P_{cp}} + N_s\right)
\cdot \Big(& c_{\mathrm{bias}} \cdot l \cdot (N' + N_s) \\
& + c_{\mathrm{feat}} \cdot d_{\mathrm{in}} \cdot \mathbb{1}[i=0] \Big),
\end{aligned}
\end{equation}
where $c_{mlp}$, $c_{attn}$, $c_{bias}$, and $c_{feat}$ are constant coefficients
fitted from profiling measurements,
and $\mathbb{1}[\cdot]$ denotes the indicator function.

\begin{figure}[t]
    \centering
    \includegraphics[width=\linewidth]{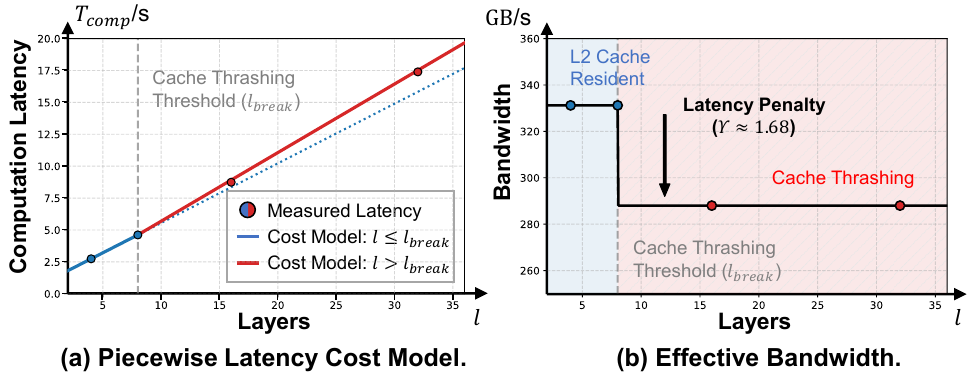}
    \vspace{-20pt}
    \caption{Validation of the piecewise latency cost model with $P_{dp}=2$.}
    \label{fig:costmodel}
\end{figure}

\textbf{Latency cost model.}
The latency cost model predicts the execution time of a single training step.
We observe that when the number of layers $l$ exceeds a threshold
$l_{break} = c_{break} / (b \times d)$,
where $c_{break}$ is a hardware-specific cache capacity constant,
the working set surpasses L2 cache capacity,
incurring an approximately $\gamma$-fold latency penalty on global memory access.
The computation latency is therefore modeled piecewise:
\begin{equation}
    T_{comp} =
    \begin{cases}
    l \times (t_{fwd} + t_{bwd}) + T_{extra} & \text{if } l \le l_{break} \\
    l \times \gamma \times (t_{fwd} + t_{bwd}) + T_{extra} & \text{if } l > l_{break}
    \end{cases},
    \label{eq:tcomp}
\end{equation}
where $T_{extra} = t_{emb} \cdot \mathbb{1}[i=0] + t_{proj} \cdot \mathbb{1}[i=P_{pp}-1]$,
and $\gamma$ is fitted from hardware profiling measurements.
Fig.~\ref{fig:costmodel} validates that
when the number of layers exceeds $l_{break}$,
cache thrashing incurs an approximately $\gamma$-fold latency increase.

Communication latency is also modeled based on profiling data.
Under the GPipe schedule, the local batch $b$ is divided into $K$ micro-batches.
The pipeline cycle time $t_{cycle}$ is determined by the bottleneck stage:
\begin{equation}
    t_{cycle} = \max_{i \in [0, P_{pp})} \frac{T_{comp}^{(i)}}{K} + t_{p2p}(S_{act}),
    \label{eq:tcycle}
\end{equation}
where $S_{act} = (b/K) \times (N'/P_{cp} + N_s) \times d$ is the boundary activation tensor size
transferred between pipeline stages,
and $T_{comp}^{(i)}$ denotes the full-batch computation time for stage $i$,
evaluated via Eq.~\ref{eq:tcomp} with the layer count $l_i$ assigned to that stage,
where dividing by $K$ gives the per-micro-batch computation time.
The data loading stall time $T_{stall}$, caused by CPU-side subgraph sampling
and attention bias construction, is measured directly during profiling.
The final single-step training latency combines the pipeline execution time,
data loading stalls, and All-Reduce over the total parameter size $S_{param}$:
\begin{equation}
    T = (K + P_{pp} - 1) \times t_{cycle}
    + T_{stall} + t_{allreduce}(S_{param}).
    \label{eq:time}
\end{equation}

\begin{algorithm}[t]
\small
\caption{Automatic Optimal Strategy Search}
\label{alg:auto-search}
\renewcommand{\algorithmicrequire}{\textbf{Input:}}
\renewcommand{\algorithmicensure}{\textbf{Output:}}
\begin{algorithmic}[1]
    \REQUIRE $W, L, M_{limit}, K$
    \ENSURE $(P_{pp}^*, P_{cp}^*, P_{dp}^*)$ and layer partition

    \STATE $T_{min} \leftarrow \infty$, BestConfig $\leftarrow \emptyset$
    \FOR{each valid $(P_{pp}, P_{cp})$ with $W = P_{pp} P_{cp} P_{dp}$}
        \STATE $dp[0][0] \leftarrow 0$; other entries $\leftarrow \infty$
        \FOR{$i = 1$ to $P_{pp}$}
            \FOR{$j = 0$ to $L-1$; $k = j+1$ to $L$}
                \IF{$\text{MemCost}(j{+}1, k) > M_{limit}$}
                    \STATE \textbf{continue}
                \ENDIF
                \STATE $T_{s} \leftarrow \text{TimeCost}(j{+}1, k)$
                \STATE $dp[i][k] \leftarrow
                \min\big(dp[i][k],\; \max(dp[i{-}1][j], T_{s})\big)$
            \ENDFOR
        \ENDFOR

        \IF{$dp[P_{pp}][L] \neq \infty$}
            \STATE $T_{step} \leftarrow
            \text{StepLatency}(dp[P_{pp}][L], P_{pp}, K)$
            \IF{$T_{step} < T_{min}$}
                \STATE BestConfig $\leftarrow
                (P_{pp}, P_{cp}, P_{dp},$ Backtrack$(dp))$
            \ENDIF
        \ENDIF
    \ENDFOR
    \RETURN BestConfig
\end{algorithmic}
\end{algorithm}

\subsubsection{Optimal Strategy Search}

The search minimizes end-to-end latency to maximize throughput.
Let $W$ denote the total number of available GPUs.
The configuration space covers all valid parallelism combinations
satisfying $W = P_{pp} \times P_{cp} \times P_{dp}$
and the heterogeneous layer partitions for each $P_{pp}$.
As discussed above, uniform partitioning causes stage imbalance,
so the search must explore heterogeneous layer assignments.

We design a dynamic programming algorithm with memory-aware pruning,
as detailed in Alg.~\ref{alg:auto-search}, where $L$ denotes the total number of model layers..
For each $(P_{pp}, P_{cp}, P_{dp})$,
the algorithm finds the optimal layer partition that balances execution time across stages.
State $dp[i][k]$ represents the minimum latency
for allocating $k$ layers into $i$ stages:
\begin{equation}
    dp[i][k] = \min_{0 \le j < k}
    \max \Big( dp[i-1][j], \; T_{comp}(j+1, k, P_{cp}) \Big),
\end{equation}
where $T_{comp}$ is computed via Eq.~\ref{eq:tcomp}.
Any branch where the predicted stage memory $M_{stage}$ exceeds the device limit $M_{limit}$
is pruned, preventing OOM errors while substantially reducing the search space.
Upon completion, the algorithm evaluates the final step latency $T_{step}$
via Eq.~\ref{eq:time} and outputs the optimal configuration
$(P_{pp}^*, P_{cp}^*, P_{dp}^*)$ with its layer partition.
\section{Experiment}

\subsection{Experimental Setup}
\label{sec:setup}

\textbf{Datasets.}
We use five public graph datasets.
End-to-end experiments use Products as the large-scale benchmark,
Reddit and Reddit2 as medium-scale datasets, and Citeseer for small-scale evaluation.
The Computers dataset serves as the ablation and configuration search benchmark,
since it is large enough for all parallelism strategies
to exhibit meaningful trade-offs, yet tractable for full configuration search.
The Arxiv dataset provides additional cross-architecture accuracy validation.

\textbf{Baselines.}
We compare against two baselines.
ANS-GT~\cite{zhang2022hierarchical} is a hierarchical GT training framework
that supports single-GPU training only.
TorchGT~\cite{Zhang2024torchgt} is based on Graphormer~\cite{ying2021transformers},
supporting multi-GPU sequence-parallel full attention.
MegaGraph integrates the same hierarchical GT model as ANS-GT,
extended with our hybrid parallelism backend and automatic search engine.
The comparison with ANS-GT isolates system-level gains under the same model,
while TorchGT provides a cross-architecture accuracy reference.

\textbf{End-to-End Experiment Configuration.}
For the end-to-end comparison, all three systems are configured identically:
16 layers, 8 heads, hidden dimension of 64, and FFN dimension of 64.
MegaGraph and ANS-GT share the same hierarchical GT model
with attention bias dimension of 4 and 1 global node.
TorchGT is based on Graphormer with its default structural encoding,
and runs in full attention mode to align the attention scope.
The batch size is 1024 for small and medium graphs, and 8192 for large graphs.
TorchGT on a single GPU goes OOM with batch size 8192,
so we reduce its single-GPU batch size to 6800, denoted TorchGT-1G$'$.
Throughout end-to-end comparisons, the suffixes \textbf{-1G} and \textbf{-2G} denote single-GPU and two-GPU training configurations, respectively.
All experiments train for 100 epochs
with peak learning rate 2e-4 and weight decay 0.01.

\textbf{Hardware and Metrics.}
End-to-end and ablation experiments run on dual A100 (80 GB),
cost model validation on dual Tesla M40 (12 GB),
and Pareto search on 4$\times$RTX 4090 (24 GB).
We measure test accuracy (\%), steady-state GPU training time (s/epoch),
and peak GPU memory (GB).

\subsection{End-to-End Performance}
\label{sec:endtoend}

\begin{figure}[t]
    \centering
    \includegraphics[width=1\columnwidth]{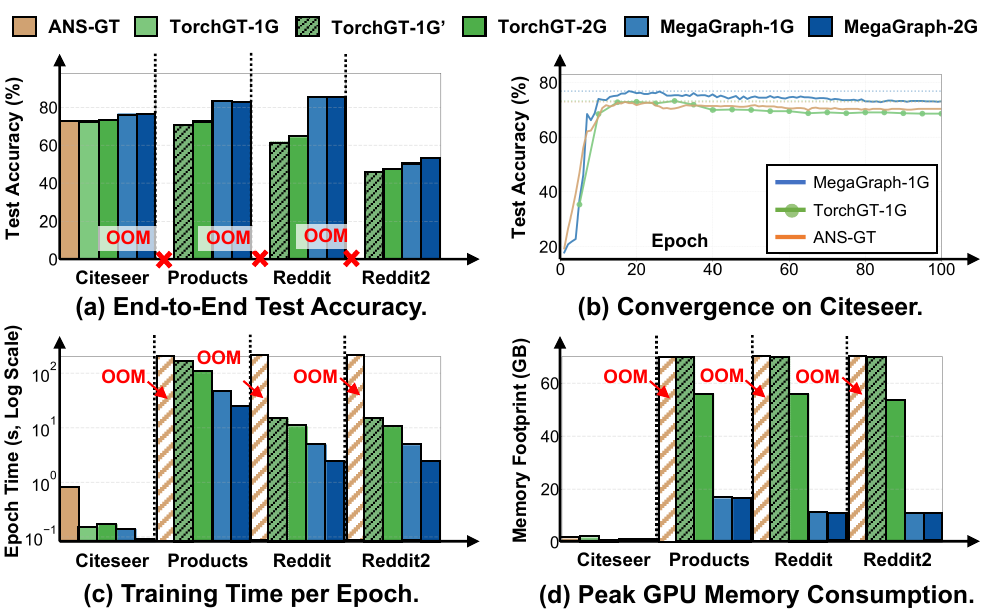}
    \vspace{-10pt}
    \caption{End-to-end performance comparison using a 16-layer, 8-head GT with hidden and FFN dimension 64. OOM entries indicate training failure. TorchGT-1G$'$ denotes a reduced batch size of 6800.}
    \label{fig:endtoend}
\end{figure}

\textbf{Accuracy Preservation \& Convergence.}
We first verify that the parallel backend preserves model quality.
MegaGraph single-GPU and two-GPU results differ by less than 0.2 percentage points across all datasets.
As shown in Fig.~\ref{fig:endtoend}(a),
MegaGraph consistently outperforms TorchGT across all evaluated datasets,
with the advantage being particularly pronounced on large-scale graphs,
where MegaGraph exceeds TorchGT by over 10 percentage points on Products.
This reflects the architectural difference between hierarchical GT and flat Graphormer:
on small and medium graphs where ANS-GT can also run,
MegaGraph and ANS-GT achieve comparable accuracy,
confirming that the parallel backend introduces no accuracy degradation.
We also trained ANS-GT and TorchGT on Arxiv under the same model configuration,
where ANS-GT similarly outperforms TorchGT by a large margin,
further corroborating this conclusion.
Fig.~\ref{fig:endtoend}(b) shows the convergence curves on Citeseer,
where MegaGraph achieves accuracy comparable to ANS-GT,
confirming that the parallel backend introduces no accuracy degradation,
while both hierarchical GT models consistently outperform TorchGT.

\textbf{Memory Elimination \& Speedup.}
ANS-GT, sharing the same hierarchical GT model,
goes OOM on both Products and Reddit.
MegaGraph's hybrid parallelism backend eliminates this bottleneck,
enabling training on these large graphs.
Scaling from one to two GPUs, MegaGraph achieves near-linear speedup
(1.88$\times$ on Products, 2.02$\times$ on Reddit),
and delivers 4.51$\times$ and 4.36$\times$ speedup over TorchGT on the two-GPU setting,
as shown in Fig.~\ref{fig:endtoend}(c).
On memory, MegaGraph-1G requires only 16.53~GB on Products,
a 77.8\% reduction from TorchGT-1G$'$'s 74.4~GB
even after TorchGT reduces its batch size to 6800 to avoid OOM,
and MegaGraph-2G achieves a further 70.5\% reduction over TorchGT-2G,
as shown in Fig.~\ref{fig:endtoend}(d).

\vspace{-6pt}
\subsection{Accuracy of Cost Model}
\label{sec:costmodel}

\begin{figure}[t]
    \centering
    \includegraphics[width=1\columnwidth]{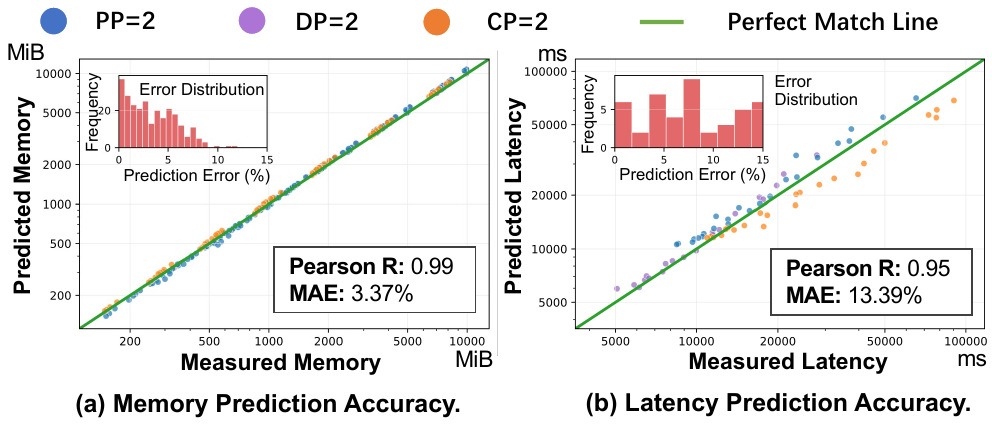}
    \vspace{-10pt}
    \caption{Accuracy of the memory and latency cost models. Diagonal lines indicate perfect prediction ($Y=X$).}
    \label{fig:predict}
\end{figure}

As shown in Fig.~\ref{fig:predict},
we validated the cost models on dual Tesla M40 across 216 configurations
(covering $b \in \{1,2,4,8\}$, $l \in \{2,4,8,16,24,32\}$,
$h \in \{4,8,16\}$, $d \in \{32,64,128\}$).
Memory prediction achieves MAPE = 3.37\% with R$^2$ = 0.994,
which is sufficient for the search engine's memory-aware pruning.
When layers exceed $l_{break} \approx 8$,
the working set surpasses L2 cache capacity,
causing a $\gamma \approx 1.68\times$ latency penalty.
The piecewise linear model accurately captures this transition (R$^2$ = 0.961).
Model parameters are fitted from profiling measurements,
naturally supporting cross-GPU-architecture transfer.
The communication terms $t_{p2p}$ and $t_{allreduce}$ are directly measured
from the target hardware rather than analytically modeled,
and their accuracy is further corroborated by the end-to-end Pareto search results
in Section~\ref{sec:search},
where the search engine's recommendations consistently match observed performance.

\subsection{Automatic Configuration Search}
\label{sec:search}

\textbf{Ablation Study.}
We isolate each parallelism strategy's contribution on Computers dataset
with 32 layers, hidden dimension of 128, 32 heads, batch size 256, on 2$\times$A100.
This ablation validates the two core design choices of our parallelism backend.
Graph-aware context parallelism (GA-CP) targets the per-layer attention memory bottleneck
arising from the $N \times N$ attention score and bias matrices in GT training.
Heterogeneous pipeline parallelism addresses the workload imbalance
caused by the graph embedding table and data construction overhead
concentrated in the first pipeline stage.
The unbalanced pipeline parallelism row further validates
the necessity of automatic layer partitioning.

\begin{table}[h]
    \centering
    \caption{Ablation of parallelism strategies. Each configuration is denoted as a $(P_{pp}, P_{cp}, P_{dp})$ triple. Mem$_0$ and Mem$_1$ report peak memory per GPU.}
    \label{tab:ablation}
    \vspace{-2pt}
    \footnotesize
    \renewcommand{\arraystretch}{1.3}
    \begin{tabular}{|>{\centering}m{1.4cm}|c|c|c|c|}
        \hline
        \textbf{Config} & \textbf{Time/ep} & \textbf{Mem$_0$} & \textbf{Mem$_1$} & \textbf{Strategy} \\ \hline
        (1,1,1) Baseline & 8.13s & 70.56 GB & N/A & None \\ \hline
        (1,2,1) CP=2 & 4.92s & \textbf{13.50} GB & 13.50 GB & GA-CP \\ \hline
        (2,1,1) PP=2 balanced & \textbf{4.35s} & 36.83 GB & 34.06 GB & Hetero.\ PP \\ \hline
        (2,1,1) PP=2 unbalanced & 6.00s & 53.68 GB & 17.21 GB & Naive PP \\ \hline
    \end{tabular}
    \vspace{-3pt}
\end{table}

As shown in Tab.~\ref{tab:ablation},
enabling graph-aware context parallelism allows each GPU
to hold only half of the attention sequence,
reducing peak memory from 70.56~GB to 13.50~GB, a 5.2$\times$ reduction,
with only 13\% slowdown compared to the best pipeline parallelism configuration.
With naive even-split layer partitioning in pipeline parallelism,
the embedding overhead causes the bottleneck GPU to reach 53.68~GB,
strictly dominated by context parallelism in both speed and memory,
validating the necessity of heterogeneous layer partitioning.
Heterogeneous pipeline parallelism automatically adjusts
the layer assignment across stages,
achieving the best throughput at 4.35~s/epoch,
13\% faster than context parallelism,
but with peak memory at 36.83~GB, 2.7$\times$ that of context parallelism.
Neither strategy dominates across both metrics simultaneously,
confirming that no single parallelism choice is universally optimal
and that an automatic search engine is essential for selecting
the right configuration under different hardware and memory constraints.

\begin{figure}[t]
    \centering
    \includegraphics[width=1\columnwidth]{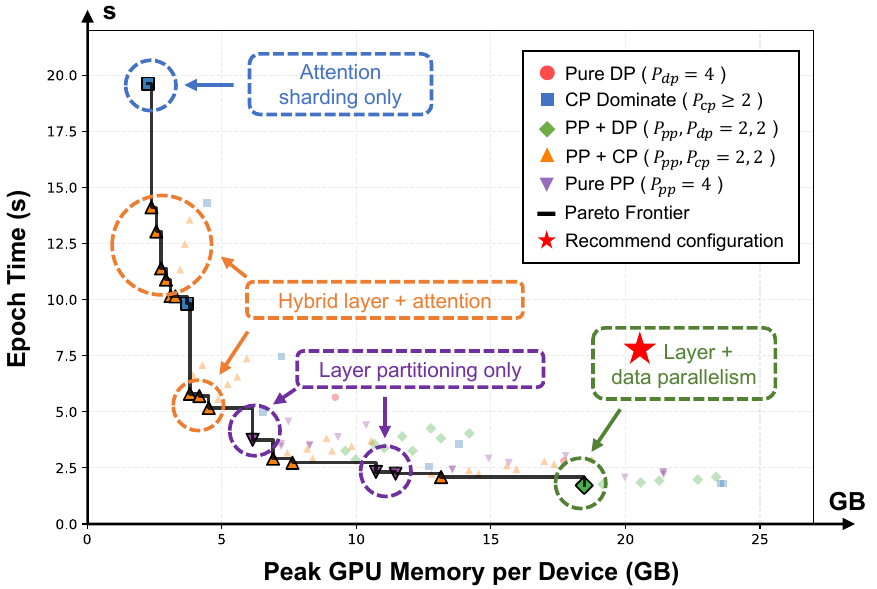}
    \vspace{-12pt}
    \caption{Pareto-optimal configuration search on a Graph Transformer model, with 32 layers across 166 configurations on 4$\times$RTX 4090.}
    \label{fig:pareto}
\end{figure}

\textbf{Pareto Frontier.}
Fig.~\ref{fig:pareto} shows the Pareto frontier across 166 configurations
in 5 strategy regions on 4$\times$RTX 4090.
As shown in the figure, each strategy occupies a distinct region of the frontier.
DP dominates on throughput but does not reduce per-GPU memory,
while CP achieves the lowest memory at the cost of communication overhead.
PP sits between the two, and hybrid configurations cover the remaining trade-offs.
None of the Pure DP configurations lie on the Pareto frontier,
as CP+DP hybrid strategies match their throughput with lower memory,
confirming that the optimal strategy is hardware- and objective-dependent.
The DP algorithm with memory pruning finds the global optimum in $\sim$3s.
Under the 24~GB RTX 4090 constraint, the search engine selects a PP+DP hybrid, with all recommended configurations fitting within device memory.

\section{Conclusion}

To address the memory and computational bottlenecks
in training large-scale Graph Transformers,
this paper proposes MegaGraph,
an automated hybrid parallel training framework.
MegaGraph designs three specialized parallelism strategies:
graph-aware context parallelism addresses the attention memory bottleneck
while preserving topological bias alignment,
heterogeneous pipeline parallelism compensates
for the workload imbalance caused by graph embedding and data construction overhead,
and hybrid data parallelism provides extra acceleration along the graph data dimension.
To navigate the combinatorially large configuration space these strategies induce,
an automatic search engine follows a \textit{Profile $\to$ Model $\to$ Search} workflow
with precise cost models and memory-aware pruning,
identifying the optimal parallelism configuration within seconds.

Experimental results demonstrate that MegaGraph enables training
on large-scale graphs where existing baselines fail with OOM,
reducing peak memory by 77.8\% and achieving 4.51$\times$ training speedup
compared with TorchGT, while maintaining model accuracy.
In future work, we plan to extend MegaGraph to dynamic graph architectures
and explore elastic scheduling mechanisms
on larger-scale heterogeneous computing systems.

\vspace{-8pt}

\bibliographystyle{IEEEtran}
\small
\bibliography{iccad2025}

\end{document}